\documentclass[pdflatex,sn-mathphys-num]{sn-jnl}

\usepackage{graphicx}%
\usepackage{multirow}%
\usepackage{amsmath,amssymb,amsfonts}%
\usepackage{amsthm}%
\usepackage{mathrsfs}%
\usepackage[title]{appendix}%
\usepackage{xcolor}%
\usepackage{textcomp}%
\usepackage{manyfoot}%
\usepackage{booktabs}%
\usepackage{algorithm}%
\usepackage{algorithmicx}%
\usepackage{algpseudocode}%
\usepackage{listings}%
\usepackage{amsmath}%

\theoremstyle{thmstyleone}%
\theoremstyle{thmstyletwo}%

\theoremstyle{thmstylethree}%

\begin{document}

\title[Article Title]{Structured Relevance Assessment for Robust Retrieval-Augmented Language Models}


\author*[1]{\fnm{Astitva Veer} \sur{Garg}}\email{gargveerastitva@gmail.com}
\equalcont{These authors contributed equally to this work.}

\author[2]{\fnm{Aryan} \sur{Raj}}\email{aryanraj2713@gmail.com}
\equalcont{These authors contributed equally to this work.}

\author[3]{\fnm{Dr.Anitha} \sur{D}}\email{anithad@srmist.edu.in}

\affil*[1]{\orgdiv{Department of Computational Intelligence}, \orgname{SRM Institute of Science and Technology}, \orgaddress{\street{Kattankulathur}, \city{Chennai}, \postcode{603203}, \state{Tamil Nadu}, \country{India}}}

\affil[2]{\orgdiv{Department of Computational Intelligence}, \orgname{SRM Institute of Science and Technology}, \orgaddress{\street{Kattankulathur}, \city{Chennai}, \postcode{603203}, \state{Tamil Nadu}, \country{India}}}

\affil[3]{\orgdiv{Department of Computational Intelligence}, \orgname{SRM Institute of Science and Technology}, \orgaddress{\street{Kattankulathur}, \city{Chennai}, \postcode{603203}, \state{Tamil Nadu}, \country{India}}}


\abstract{\textbf{Purpose:} This paper addresses the challenges faced by Retrieval-Augmented Language Models (RALMs) in reducing factual errors by introducing a  framework for structured relevance assessment. The aim is to enhance the robustness of RALMs by improving document relevance evaluation, balancing intrinsic and external knowledge, and managing unanswerable queries effectively.

\textbf{Methods:} We propose a multi-dimensional scoring system for document relevance, considering semantic matching and source reliability. Our approach includes embedding-based relevance scoring, the use of synthetic training data with mixed-quality documents, and specialized benchmarking on niche topics. Additionally, we implement a knowledge integration mechanism and an "unknown" response protocol to handle queries where knowledge is insufficient.

\textbf{Results:} Preliminary evaluations show significant reductions in hallucination rates and improved transparency in reasoning processes.

\textbf{Conclusion:} This work advances the development of more reliable question-answering systems capable of functioning effectively in dynamic environments with variable data quality. While challenges remain in accurately distinguishing credible information and balancing system latency with thoroughness, the proposed framework marks a step forward in enhancing the reliability of RALMs.}

\keywords{Retrieval-Augmented Language Models, Relevance Assessment, Knowledge Integration, Hallucination Reduction, Question-Answering Systems}



\maketitle

\section{Introduction}\label{sec1}
Large Language Models (LLMs) have demonstrated remarkable capabilities in natural language understanding and generation tasks, revolutionizing how machines interact with human language. Despite their impressive performance, these models continue to struggle with factual accuracy, often producing content that appears plausible but contains incorrect information—a phenomenon commonly referred to as "hallucination"\cite{Ji2023Survey}.

However, despite their conceptual elegance, RALMs face several critical challenges that undermine their effectiveness in real-world scenarios. First, these systems often struggle to distinguish between relevant and irrelevant retrieved documents, treating all retrievals with equal importance regardless of their actual utility for answering the query at hand.

Second, standard RALMs frequently over-rely on external retrievals even in situations where their intrinsic knowledge would be sufficient or more reliable. This rigid dependence on external sources fails to leverage the substantial knowledge already encoded in model parameters during pre-training and fine-tuning.

Perhaps most concerning is RALMs' inability to acknowledge knowledge gaps when confronted with queries that cannot be answered based on either retrieved information or intrinsic knowledge. Instead of transparently communicating limitations—a crucial capability for trustworthy AI systems—these models often generate fabricated responses that appear authoritative despite lacking factual foundation.

\section{Background and Related Work}\label{sec2}

\subsection{Retrieval-Augmented Generation (RAG)}\label{subsec2}
RAG systems aim to address these limitations by combining the generative capabilities of LLMs with information retrieval techniques. The basic RAG architecture typically consists of two main components:
\begin{enumerate}
    \item Knowledge Retriever: This component searches a large corpus of documents to find relevant information based on the input query.\cite{Karpukhin2020DPR}
    \item Knowledge-Augmented Encoder: This component integrates the retrieved information with the original query to generate a response.
\end{enumerate}
The RAG process can be broken down into several key stages:
\begin{enumerate}
    \item Indexing: Converting reference data into embeddings and storing them in a vector database.
    \item Retrieval: Selecting relevant documents based on the user query.
    \item Augmentation: Incorporating retrieved information into the prompt for the LLM.
    \item Generation: Producing a response based on both the query and retrieved context.
\end{enumerate}

\subsection{Advantages of RAG}\label{subsec3}
RAG offers several significant benefits over traditional LLMs:
\begin{enumerate}
    \item Improved Accuracy: By grounding responses in external knowledge, RAG reduces hallucinations and improves factual accuracy.
    \item Up-to-date Information: RAG can access and incorporate the latest information without requiring model retraining.
    \item Transparency: The retrieval step provides a natural mechanism for source attribution.
    \item Adaptability: RAG systems can be more easily adapted to new domains or specialized knowledge areas.
    
\end{enumerate}

\subsection{Advantages and Challenges of RAG}\label{subsec3}

RAG (Retrieval-Augmented Generation) offers several benefits but also comes with notable challenges:

\begin{enumerate}
\item Improved Accuracy and Transparency: By grounding responses in external knowledge, RAG reduces hallucinations, improves factual accuracy, and provides a natural mechanism for source attribution.
\item Up-to-date and Adaptable Information: RAG can access the latest information without requiring model retraining and can be easily adapted to new domains or specialized knowledge areas.
\item Retrieval Quality and Integration Complexity: Ensuring that retrieved documents are relevant and effectively integrating them with the LLM's existing knowledge remains a significant challenge.\cite{Asai2023SelfRAG}
\item Latency and Scalability: Managing the additional time required for retrieval and efficiently handling large, growing knowledge bases are critical hurdles.\cite{Johnson2019FAISS}
\end{enumerate}

\section{Our Approach}\label{sec3}
The proposed approach addresses several key limitations of current RAG implementations:
\begin{enumerate}
    \item \textbf{Multi-dimensional Scoring:} We propose a comprehensive system for evaluating document relevance that considers both semantic matching and source reliability. This goes beyond simple similarity-based retrieval to ensure that only the most pertinent and trustworthy information is used.\cite{Wang2023SelfRAG}
    \item \textbf{Balanced Knowledge Integration:} Our framework implements a sophisticated mechanism to optimally balance the LLM's intrinsic knowledge with external retrievals\cite{Asai2023SelfRAG}. This addresses the tendency of some RAG systems to over-rely on retrieved information, even when the model's existing knowledge might be more reliable.
    \item \textbf{"Unknown" Response Protocol:} We introduce clear confidence thresholds that enable the system to acknowledge when it lacks sufficient information to answer a query. This enhances transparency and reduces the risk of hallucinations or incorrect responses.
    \item \textbf{Open-Sourced Training Data:} MuskumPillerum/General-Knowledge  Dataset from Hugging Face incorporates mixed-quality documents in the training process, helping the model learn to effectively discriminate between valuable and misleading information.\cite{Wang2023SelfRAG}.
    \item \textbf{Specialized Benchmarking:} We develop benchmarks focused on niche topics to evaluate the system's performance in handling specialized knowledge domains.
\end{enumerate}

\section{Structured Relevance Assessment Framework}\label{sec4}
The Structured Relevance Assessment Framework enhances Retrieval-Augmented Language Models (RALMs) by systematically addressing key limitations: indiscriminate document retrieval, over-reliance on external sources, and hallucination risks. It employs a multi-dimensional scoring system that evaluates both semantic relevance and source reliability\cite{Gao2023ReAct}, enabling more discerning use of retrieved information.

Central to the framework is a balanced knowledge integration mechanism that dynamically weights parametric knowledge against external retrievals based on context-specific reliability assessments\cite{Wang2023SelfRAG}. This prevents over-dependence on either information source while maintaining response accuracy.

A critical innovation involves explicit "unknown" responses triggered by configurable confidence thresholds. When neither internal knowledge nor retrieved documents meet reliability standards, the system abstains from answering rather than generating potentially incorrect information.

Evaluation using niche-topic benchmarks demonstrates 40\%\cite{StructuredRLAMs} reduction in hallucination rates compared to standard RALMs, alongside improved reasoning traceability.

\section{Document Relevance Evaluation}\label{sec5}
Document Relevance Evaluation is a critical component of our proposed Structured Relevance Assessment Framework for Robust Retrieval-Augmented Language Models (RALMs). This approach addresses key limitations of traditional RALMs by implementing a sophisticated multi-dimensional scoring system to assess the credibility and relevance of retrieved documents.

\subsection{Multi-Dimensional Scoring System}\label{subsec1}
Our framework employs a comprehensive evaluation method that considers both semantic matching and source reliability. This goes beyond simple similarity-based retrieval techniques commonly used in existing systems.\cite{Karpukhin2020DPR}

\subsection{Semantic Matching}\label{subsec2}
We utilize embedding-based methods to calculate relevance scores between queries and documents. This approach allows us to capture the semantic relationships between the user's intent and the content of retrieved documents. By leveraging advanced language models, we can generate high-quality embeddings that represent the nuanced meanings of both queries and documents.\cite{Ram2023InContext}

\subsection{Source Reliability}\label{subsec3}
In addition to semantic matching, our system incorporates heuristics to evaluate source credibility. This is crucial for distinguishing between high-quality, trustworthy information and potentially misleading or unreliable sources. We implement a rating system inspired by the NID model, which classifies sources on a scale from A (highly reliable) to E (unreliable), with F reserved for cases where judgment cannot be made.

\subsection{Relevance Scoring Process}\label{subsec4}
The relevance scoring process in our framework involves several key steps:

The framework processes inputs by first generating embeddings for queries and documents using advanced language models. It then calculates semantic relevance through cosine similarity comparisons. Concurrently, document sources are evaluated against reliability criteria and historical performance data. These semantic and source assessments are combined to produce composite relevance scores, enabling informed retrieval decisions.

The relevance score $S_i$ for a document $d_i$ is computed as a weighted combination of semantic similarity and source reliability:
\[
S_i = \alpha \cdot \text{Sim}(E_Q, E_{d_i}) + \beta \cdot R(d_i)
\]
Where:
\begin{itemize}
    \item $\text{Sim}(E_Q, E_{d_i})$: Semantic similarity between the query embedding $E_Q$ and document embedding $E_{d_i}$.
    \item $R(d_i)$: Source reliability score of document $d_i$, based on predefined heuristics.
    \item $\alpha, \beta$: Weight parameters that balance the importance of semantic similarity and source reliability ($\alpha + \beta = 1$).
\end{itemize}

\subsection{Adaptive Thresholding}\label{subsec5}
Our framework implements adaptive thresholding to determine which documents are sufficiently relevant for inclusion in the response generation process. This helps mitigate the risk of incorporating irrelevant or misleading information.

Documents are ranked using a normalized scoring function to ensure comparability across metrics:
\[
S_i^{\text{norm}} = Z_1 \alpha \cdot (\text{Sim}(E_Q, E_{d_i})) + Z_2 \beta \cdot (R(d_i))
\]
Where:
\begin{itemize}
    \item $Z_1, Z_2$: Normalization factors for semantic similarity and source reliability.
\end{itemize}

\section{Knowledge Integration Protocol}\label{sec6}
This protocol aims to create a balanced approach to integrating the model's internal knowledge with external retrievals, ensuring the most reliable source is prioritized for answer generation\cite{Asai2023SelfRAG}.

The decision to integrate intrinsic knowledge or external retrievals is based on confidence thresholds. The final response $R$ is determined by:
\[
R =
\begin{cases} 
R_M, & \text{if } S_{\max} < T_R \text{ and } C_M > T_M \\
R_C, & \text{if } S_{\max} > T_R \text{ and } C_M > T_M \\
R_R, & \text{if } S_{\max} > T_R \text{ and } C_M < T_M \\
\text{"Unknown"}, & \text{if } S_{\max} < T_R \text{ and } C_M < T_M
\end{cases}
\]
Where:
\begin{itemize}
    \item $S_{\max} = \max(S_i)$: Highest relevance score among retrieved documents.
    \item $T_R, T_M$: Confidence thresholds for retrieval relevance and intrinsic model knowledge, respectively.
    \item $C_M$: Confidence score of the language model's intrinsic knowledge.
    \item $R_M, R_C, R_R$: Responses generated using intrinsic knowledge, combined knowledge (retrieved + intrinsic), or retrieved knowledge only.
\end{itemize}

\subsection{Mechanism Design}\label{subsec1}
The protocol implements a "switch" mechanism that dynamically determines whether to rely on the model's intrinsic knowledge or external retrievals based on their relative reliability for each specific query\cite{Wang2023SelfRAG}.

\subsection{Training Process}\label{subsec2}
To effectively implement this protocol, the model is trained on synthetic data that deliberately mixes noisy and high-quality ("golden") documents.

\subsection{Confidence Thresholds}\label{subsec3}
The Knowledge Integration Protocol incorporates confidence thresholds for both retrievals and internal knowledge. When neither the retrievals nor the internal knowledge meet these thresholds, the system is designed to trigger an "unknown" response, enhancing transparency and reducing the risk of hallucinations\cite{Lin2022Teaching}.

Confidence thresholds are dynamically adjusted based on query complexity and domain sensitivity. The threshold $T_x$ is defined as:
\[
T_x = T_x^{\text{base}} + f(\sigma_x)
\]
Where:
\begin{itemize}
    \item $T_x^{\text{base}}$: Base threshold value.
    \item $f(\sigma_x) = k_x \cdot (\sigma_x - 1)$: Adjustment factor based on query complexity ($k_x > 0$).
    \item $x = R, M$: Refers to retrieval or model thresholds.
\end{itemize}

\section{"Unknown" Response Capability}\label{sec7}

The probability of triggering an "Unknown" response is modeled as:
\[
P_{\text{"Unknown"}} = 1 - P(R | Q)
\]
Where:
\[
P(R | Q) = P(S_{\max} > T_R) + P(C_M > T_M) - P(S_{\max} > T_R, C_M > T_M)
\]
represents the joint probability of retrieval relevance and intrinsic model confidence exceeding their thresholds.

\section{Experimental Setup and Methodology}\label{sec8}
Our experimental setup for evaluating the Structured Relevance Assessment Framework incorporates state-of-the-art small language models and efficient vector database technologies. This approach allows us to assess the framework's performance in real-world scenarios while maintaining computational efficiency.

\subsection{Model Selection}\label{subsec1}
We utilize a range of small language models (SLMs) with parameters between 1 to 2 billion using Ollama, focusing on recent advancements in model architecture and performance:
\begin{itemize}
    \item DeepSeek-R1-1.5B: This model, with 1.5 billion parameters, represents DeepSeek's\cite{Geng2023Deepseek} first-generation reasoning model distilled from Qwen2.5. It offers a balance between computational efficiency and reasoning capabilities.
    \item Llama3.2-1B: Developed by Meta, this 1-billion-parameter model is optimized for edge devices\cite{Prakash2023OPENLAMMA}, making it suitable for testing our framework in resource-constrained environments.
    \item Qwen2.5-1.5B: Alibaba's 1.5-billion-parameter model, designed for multilingual applications, allows us to evaluate the framework's performance across different languages.
\end{itemize}

\subsection{Methodology}\label{subsec4}
Our methodology focuses on evaluating the effectiveness of our Structured Relevance Assessment Framework:
The evaluation framework constructs a diverse dataset of query-document pairs, generates embeddings for semantic analysis, and implements multi-dimensional scoring combining semantic matching with source reliability\cite{Gao2023ReAct}. It tests knowledge integration protocols\cite{Wang2023SelfRAG} and "unknown" response thresholds, measuring retrieval accuracy, response quality, latency, and abstention rates. Comparative analysis against baseline RALMs validates reduced hallucinations and improved reliability in real-world scenarios. 

\begin{algorithm}
\caption{Structured Relevance Assessment Framework for RALMs \cite{StructuredRLAMs}}\label{alg:framework}
\begin{algorithmic}[1]
\Require Query $Q$, Document Corpus $D$, Language Model $M$, Vector Database $V$
\Ensure Accurate response $R$ or "Unknown" if unanswerable

\State \textbf{Query Embedding Generation}
\State Generate query embedding $E_Q$ using $M$
\State Store $E_Q$ in $V$

\State \textbf{Document Retrieval}
\State Retrieve candidate documents $D_C$ from $D$ 

\State \textbf{Document Relevance Scoring}
\For {each document $d_i \in D_C$}
    \State Compute semantic similarity between $E_Q$ and $E_{d_i}$
    \State Evaluate source reliability
    \State Compute final relevance score $S_i$
\EndFor
\State Rank documents in descending order of $S_i$

\State \textbf{Knowledge Integration Protocol}
\If {$\max(S_i) < T_R$}
    \If {$\text{Conf}_M(Q) < T_M$}
        \State \Return "Unknown"
    \Else
        \State Generate response $R_M$ using $M$
        \State \Return $R_M$
    \EndIf
\Else
    \If {$\text{Conf}_M(Q) > T_M$}
        \State Generate response $R_C$ using $M$ and retrieved documents
        \State \Return $R_C$
    \Else
        \State Generate response $R_R$ using retrieved documents only
        \State \Return $R_R$
    \EndIf
\EndIf

\State \textbf{Unknown Response Protocol}
\If {No valid response is generated}
    \State Log query and \Return "Unknown"
\EndIf

\end{algorithmic}
\end{algorithm}

\section{Results and Performance Analysis}\label{sec9}

\subsection{Accuracy Comparison}\label{subsec1}
The framework achieved 100\% accuracy in identifying training/RAG data sources and 57.1\% hallucination detection, enabled by multi-dimensional relevance scoring and adaptive knowledge integration\cite{StructuredRLAMs}. It maintained similar to base response latency through optimized retrieval protocols when compared baseline models\cite{StructuredRLAMs}.

\begin{table}[h] 
\caption{Evaluation Scores Across Models\cite{StructuredRLAMs}}\label{tab:scores}%
\begin{tabular}{@{}lllll@{}} 
\toprule Model & RAG\_Dataset & Training\_Dataset & Hallucinations & Latency(s)\\
\midrule DeepSeek-R1-1.5B & 0.80 & 0.42 & 0.14 & 4.76\\ 
Llama3.2-1B & 1.00 & 0.85 & 0.85 & 6.20\\ 
Qwen2.5-1.5 & 1.00 & 1.00 & 0.28 & 7.37\\ 
\botrule 
\end{tabular}
\end{table}

The total latency $L_{\text{total}}$ for response generation is minimized by balancing retrieval time ($L_r$) and integration time ($L_i$):
\[
L_{\text{total}} = L_r + L_i
\]
Subject to:
\[
L_r = O(n), \quad L_i = O(k), \quad n > k
\]
Where:
\begin{itemize}
    \item $n, k$: Number of candidate documents retrieved and integrated.
\end{itemize}

\section{Conclusion}\label{sec10}
The proposed Structured Relevance Assessment Framework for Robust Retrieval-Augmented Language Models addresses critical challenges faced by current RALMs. By implementing a systematic method to evaluate document relevance and credibility, the framework aims to reduce noise-driven errors and enhance the overall reliability of these systems. The integration of a balanced knowledge mechanism, which dynamically prioritizes between intrinsic model knowledge and external retrievals, promises to improve the accuracy and trustworthiness of generated responses.\cite{StructuredRLAMs}

A key innovation is the introduction of an "unknown" response protocol, enabling the system to transparently reject unanswerable queries. This feature significantly reduces the risk of hallucinations and incorrect answers, a common pitfall in existing models. While challenges persist, particularly in maintaining efficiency at scale and adapting to diverse domains, the framework represents a significant step towards more robust and transparent question-answering systems.

The potential impact of this research extends to various real-world applications, from legal and medical domains to dynamic news environments. By advancing the development of more reliable and adaptable RALMs, this work contributes to the broader goal of creating AI systems that can function effectively in complex, real-world scenarios where data quality and query types are highly variable.\cite{StructuredRLAMs}

\newpage
\bibliography{sn-bibliography}
\end{document}